\documentclass[letterpaper, 10 pt, conference]{ieeeconf}  

\IEEEoverridecommandlockouts                              

\newcommand{\insertfigure}[4]{ 
	\begin{figure}[!htbp]
		\begin{center}
			\includegraphics[width=#4\textwidth]{#1}
		\end{center}
		\vspace{-0.4cm}
		\caption{#2}
		\label{#3}
	\end{figure}
}

\newcommand{\insertfigureb}[4]{ 
	\begin{figure}[b]
		\begin{center}
			\includegraphics[width=#4\textwidth]{#1}
		\end{center}
		\vspace{-0.4cm}
		\caption{#2}
		\label{#3}
	\end{figure}
}

\usepackage{graphics} 
\usepackage{epsfig} 
\usepackage{mathptmx} 
\usepackage{times} 
\usepackage{amsmath} 
\usepackage{amssymb}  

\newcommand{\comment}[1]{}
\usepackage{subcaption}
\usepackage{multirow}
\usepackage{multicol}
\let\labelindent\relax
\usepackage{enumitem}
\setlist{leftmargin=1mm}
\setlist[itemize]{leftmargin=5.5mm}
\setlist[description]{font=\normalfont\itshape}
\usepackage[normalem]{ulem}
\useunder{\uline}{\ul}{}
\usepackage{booktabs}
\usepackage{siunitx}

\title{\LARGE \bf
3D Object Instance Recognition and Pose Estimation Using Triplet Loss with Dynamic Margin
}

\author{Sergey Zakharov$^{1,2}$, Wadim Kehl$^{1}$, Benjamin Planche$^{2}$, Andreas Hutter$^{2}$, Slobodan Ilic$^{1,2}$
\thanks{$^{1}$Technical University of Munich, Munich, Germany. \newline{\tt\scriptsize \mbox{\qquad} sergey.zakharov@tum.de}, {\tt\scriptsize {\{kehl, ilics\}@in.tum.de}}}%
\thanks{$^{2}$Siemens AG, Munich, Germany
\newline{\tt\scriptsize \mbox{\qquad} \{benjamin.planche, andreas.hutter\}@siemens.com}}%
}

\begin{document}

\maketitle
\thispagestyle{empty}
\pagestyle{empty}

\begin{abstract}


In this paper, we address the problem of 3D object instance recognition and pose estimation of localized objects in cluttered environments using convolutional neural networks. Inspired by the descriptor learning approach of Wohlhart et al. \cite{wohlhart2015learning}, we propose a method that introduces the dynamic margin in the manifold learning triplet loss function. Such a loss function is designed to map images of different objects under different poses to a lower-dimensional, similarity-preserving descriptor space on which efficient nearest neighbor search algorithms can be applied. Introducing the dynamic margin allows for faster training times and better accuracy of the resulting low-dimensional manifolds. Furthermore, we contribute the following: adding in-plane rotations (ignored by the baseline method) to the training, proposing new background noise types that help to better mimic realistic scenarios and improve accuracy with respect to clutter, adding surface normals as another powerful image modality representing an object surface leading to better performance than merely depth, and finally implementing an efficient online batch generation that allows for better variability during the training phase.


We perform an exhaustive evaluation to demonstrate the effects of our contributions. Additionally, we assess the performance of the algorithm on the large BigBIRD dataset \cite{singh2014bigbird} to demonstrate good scalability properties of the pipeline with respect to the number of models.

\end{abstract}
\section{INTRODUCTION}

3D pose estimation and object instance recognition are very well known problems in computer vision. They have various applications in the fields of robotics and augmented reality. Despite their popularity, there is still a large room for improvement. The current methods often struggle from clutter and occlusions and are sensitive to background and illumination changes. Moreover, the most common pose estimation methods use a single classifier per object, making their complexity grow linearly with the number of objects for which the pose has to be estimated. For industrial purposes, however, scalable methods that work with many different objects are often desirable.

The recent advances in object instance recognition can be examined by taking a glimpse at the field of 3D object retrieval, where the goal is to retrieve similar objects from a large database given some query object. 3D retrieval methods are mainly divided into two classes based on their used object representation: model-based \cite{maturana2015voxnet} and view-based \cite{su2015multi}. The first class works on 3D models directly, whereas the latter works on 2D views of the objects making it more suitable for practical applications that work with \mbox{RGB-D} images. Depending on the object representation, 2D or 3D features are used to compare different objects. Most recent approaches in this field learn features by task-specific supervision using convolutional neural networks (CNNs). Such features outperform available handcrafted features that had dominated in previous years. The view-based models currently show state-of-the-art performance, benefiting from very fast 2D convolutions (as opposed to 3D), advances in the field of image processing, and a richer object representation. However, due to their retrieval nature, these approaches do not consider clutter typical for real scenes and mainly concentrate on retrieving the object class rather than the instance.


In this work, we study the problems of 3D pose estimation and object instance recognition to propose an efficient view-based solution inspired by the recent work of Paul Wohlhart and Vincent Lepetit \cite{wohlhart2015learning}. The authors of \cite{wohlhart2015learning} tackle both pose estimation and object instance recognition of already-detected objects simultaneously by learning a discriminative feature space using CNNs. Particularly, given a single \mbox{RGB-D} image patch containing an already-detected object in the center surrounded with the cluttered background, the descriptor CNN is used to map this patch to a lower-dimensional manifold of the computed descriptors. This manifold preserves two important properties: the large Euclidean distance between the descriptors of dissimilar objects, and the distance between the descriptors of the objects from the same class is relative to their poses. Once the mapping is learned, efficient and scalable nearest neighbor search methods can be applied on the descriptor space to retrieve the closest neighbors for which the poses and identities are known. This allows us to efficiently handle a large number of objects together with their view poses, resolving the scalability issue.

The manifold learning in \cite{wohlhart2015learning} is performed using the triplet loss function, where the triplet is a group of samples $(s_i,s_j,s_k)$ selected such that $s_i$ and $s_j$ represent similar views of the same object and $s_k$ comes either from the same object with a slightly different pose or from a completely different object. The fixed margin term in the triplet loss sets the minimum ratio for the Euclidean distance between descriptors of similar and dissimilar sample pairs. Using a fixed margin throughout the training results in a slow separation of the manifolds for different objects and similar objects with different poses, causing long training times and limited accuracy in case of short-sized descriptors. To overcome this problem, we introduce a dynamic margin in the loss function by explicitly setting the margin term as a function of an angular difference between the poses for the same object and to a constant value that is larger than the maximal possible angular difference in case of different objects. This allows faster training and better quality of the resulting lower-dimensional manifolds, which, in turn, enables the use of smaller-sized descriptors with no loss of accuracy.


Apart from this main contribution, we propose the following additions to the baseline method to improve robustness to clutter and usability in real-world scenarios:
\begin{itemize}
	\item adding in-plane rotations existing in real-world scenarios and ignored by the initial method;
	\item introducing surface normals as a powerful image modality representing the object surface; normals are calculated based on the depth map images and require no additional sensor data;
	\item introducing new background noise types for synthetic patches that help to better mimic realistic scenarios and allow for better performance when no real data is used in training;
	\item implementing the efficient online batch generation that enables better variability during the training phase by filling a different background noise for synthetic patches at every solver step.
\end{itemize}

In the evaluation section, we validate the method to demonstrate the importance of the newly introduced improvements. In addition, the performance of the algorithm on a larger dataset is evaluated, proving its good scalability with respect to the number of models. For that purpose, we have adapted the BigBIRD dataset \cite{singh2014bigbird}, which has many more models available compared to the LineMOD dataset \cite{hinterstoisser2012model} initially used.

\section{Related Work}


The great increase in the number of freely available 3D models gave rise to the methods allowing for a search in large 3D object databases. They are usually referred to as 3D retrieval methods since their task is to retrieve similar objects given some 3D object query. Our method is closely related and can be looked upon as a representative of 3D retrieval methods. However, in 3D object retrieval, the queries are taken out of the real-scene context and are thus completely clutter- and occlusion-free. Moreover, for retrieval tasks, it is not usually necessary to estimate the pose of the object that is crucial for the other applications, like grasping in robotics. Finally, typical 3D retrieval benchmarks aim to retrieve the object class rather than the instance which limits us to using datasets for the object instance recognition. Since our approach is following manifold learning approaches, we will also review the most closely related works of this domain.  

3D retrieval methods are mainly divided into two classes \cite{icke2004content}: model-based and view-based. Model-based methods work with 3D models directly, trying to represent them using different kinds of features. View-based methods, on the other hand, work with 2D views of the objects. Therefore, they do not require explicit 3D object models, making them more suitable for practical applications. Moreover, view-based methods benefit from working with 2D images, which makes it possible to use dozens of efficient methods from the field of image processing. In the past, there has been a huge amount of literature related to designing features suitable for this task~\cite{knopp2010hough,ohbuchi2008ranking,daras20103d}.

More recent approaches learn features using deep neural networks, most commonly CNNs. The reason for this is that the features learned by task-specific supervision using CNNs show better performance than handcrafted ones \cite{krizhevsky2012imagenet,girshick2014rich}. Some of the popular model-based methods, such as ShapeNet \cite{wu20153d} and VoxNet \cite{maturana2015voxnet}, take 3D binary voxel grids as an input for a 3D CNN and output a class of the object. These methods show outstanding performance and are considered state-of-the-art model-based methods. However, it was shown that even the latest volumetric model-based methods are outperformed by CNN-based multi-view approaches, e.g. by the method of Hang Su et al. \cite{su2015multi}. FusionNet~\cite{hegdefusionnet} proposes fusing information from two networks, one operating on volumetric shape representation and the other using model projections represented with depth images. It seems that 2D projections capture local spatial correlations, while voxels capture long range spatial correlations. Our algorithm falls into the group of view-based methods, but instead of defining the class of the object, a specific instance is given as an output. Moreover, robustness to the background clutter is necessary since we test on real-scene scenarios.  

Another aspect that is closely related to our work is manifold learning \cite{pless2009survey}. Manifold learning is an approach to non-linear dimensionality reduction, motivated by the idea that high-dimensional data, e.g. images, can be efficiently represented in a lower-dimensional space. This concept using CNNs is well studied in \cite{hadsell2006dimensionality}. In order to learn the mapping, they use the so called Siamese network, which takes two inputs instead of one, and a specific cost function. The cost function is defined such that the squared Euclidean distance between similar objects is minimized and for dissimilar object the hinge loss is applied forcing them to be pulled apart using a margin term. In the article, they apply this concept to a problem of pose estimation. 
The work \cite{masci2014multimodal} extends this idea even further. They present a framework for multi-modal similarity-preserving hashing where an object coming from single or multiple modalities, e.g. text and images, is mapped to another space where similar objects are mapped as close together as possible and dissimilar as distant as possible.

The latest manifold learning approaches use recently introduced triplet networks, which outperform Siamese networks in generating well separated manifolds \cite{hoffer2015deep}. A triplet network, as the name suggests, takes three images as an input (instead of two in the case of Siamese networks), where two images belong to the same class and the third one to another class. The cost function tries to map the output descriptors of the images of the same class closer together than those of a different class. This allows for faster and more robust manifold learning since both positive and negative samples are taken into account within a single term. The method, recently proposed by Paul Wohlhart and Vincent Lepetit \cite{wohlhart2015learning} and inspired by these recent advances, maps the input image data directly to the similarity-preserving descriptor space using a triplet CNN with a specifically designed loss function. The loss function imposes two constrains: the Euclidean distance between the views of dissimilar objects is large, whereas the distance between the views of objects of the same class is relative to their poses. Therefore, the method learns the embedding of the object views into a low-dimensional descriptor space. Object instance recognition is then resolved by applying efficient and scalable nearest neighbor search methods on the descriptor space to retrieve the closest neighbors. Moreover, apart from only finding the object's pose, it also finds its identity, solving two separate problems at the same time and further increasing the value of this method. The approach of~\cite{guo2016multi} adds classification loss to the triplet loss and learns the embedding from the input image space to a discriminative feature space. This approach is tailored to the object class retrieval task and performs training using real images only, not rendered 3D object models.

In this paper, we follow the approach of \cite{wohlhart2015learning} and improve it in various ways, first by introducing dynamic margin in the loss function, allowing for faster training and shorter descriptors, then by making it rotation-invariant by learning in-plane rotations, including surface normals as strong and complementing modality to \mbox{RGB-D} data and providing better modeling of the background, making it more robust to the background clutter.

\section{Method}


In this section, the theoretical and implementation details of the method are presented. More specifically, such aspects as dataset generation, loss function, and improvements of the method are addressed.
\subsection{Dataset Generation}
\label{sec:dataset_generation}
The datasets we use contain the following data: 3D mesh models of the objects and \mbox{RGB-D} images of the objects in real environments with their camera poses. Using these data, we generate three sets: the training set $S_{train}$, the template set $S_{db}$ and the test set $S_{test}$. The training set is used exclusively for the purpose of training the network. The test set $S_{test}$, as its name suggests, is used only in the test phase for evaluation. The template set $S_{db}$ is used in both training and test phases. Each set consists of samples, where each sample $s = (x, c, q)$ is made of an image $x$, the identity of the object $c$, and the pose $q$.


The first step in preparing the data is the generation of samples for the sets. Our sets are constructed from two types of imaging data: real and synthetic. The real images represent the objects in real-world environments and are generated using a commodity \mbox{RGB-D} sensor, e.g. Kinect or Primesense. They have to be provided together with the dataset. The synthetic images, however, are not initially available and must be  generated by rendering provided textured 3D mesh models.

\begin{figure}[!htbp]
	\centering
	\begin{subfigure}[b]{.5\linewidth}
		\centering
		\includegraphics[width=\linewidth]{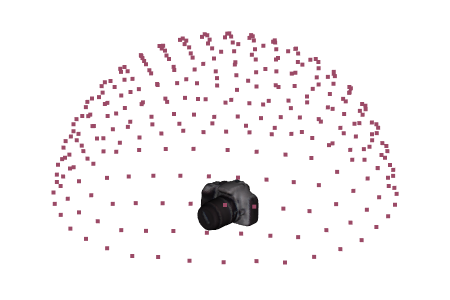}
		\caption{Template set sampling}
		\label{fig:coarse}
	\end{subfigure}%
	\begin{subfigure}[b]{.5\linewidth}
		\centering
		\includegraphics[width=\linewidth]{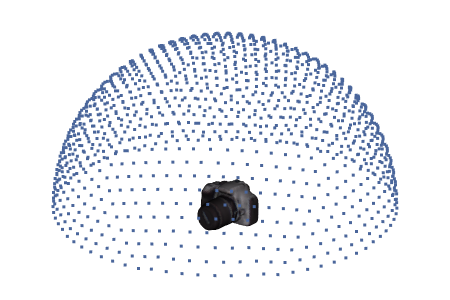}
		\caption{Training set sampling}
		\label{fig:fine}
	\end{subfigure}
	\caption{Different sampling types: each vertex represents a camera position from which the object is rendered.}
	\label{fig:samplings}
\end{figure}

Given 3D models of the objects, we render them from different view points covering the upper part of the object in order to generate synthetic images. In order to define the rendering views, an imaginary icosahedron is placed on top of the object, where each vertex defines a camera position. To make the sampling finer, each triangle is recursively subdivided into four triangles. The method defines two different sampling types: a coarse one (Fig.~\ref{fig:coarse}), achieved by two subdivisions of the icosahedron, and a fine one (Fig.~\ref{fig:fine}), achieved by three consecutive subdivisions. The coarse sampling is used to generate the template set $S_{db}$, whereas the fine sampling is used for the training set $S_{train}$. For each camera pose (vertex) an object is rendered on an empty (black) background and both RGB and depth channels are stored.

\insertfigureb{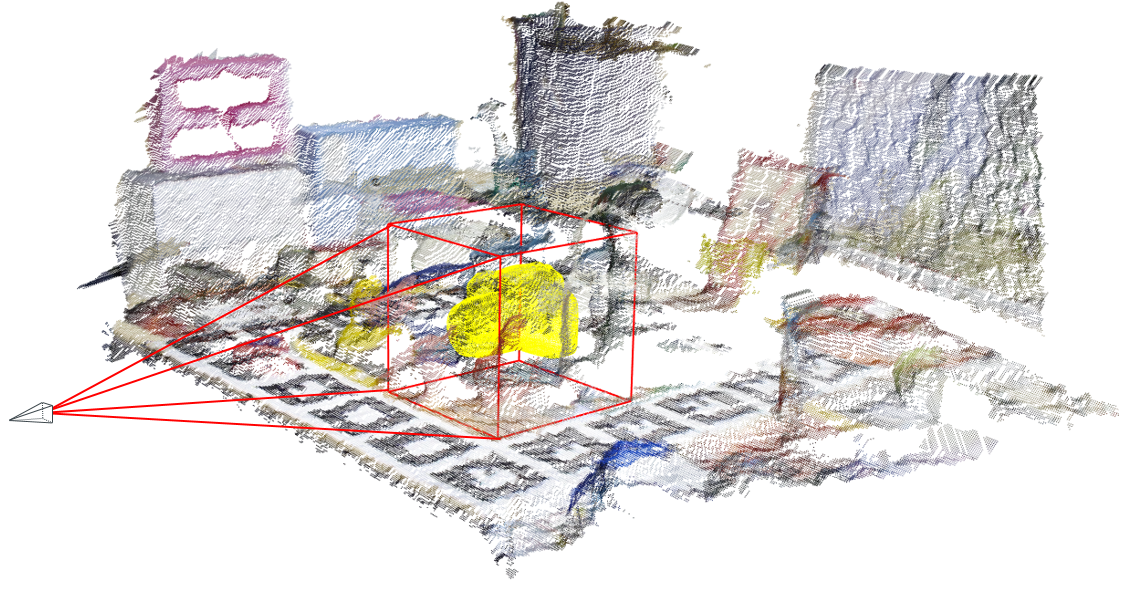}{Patch extraction: the object of interest (shown in yellow) is covered by the cube of 40 \si{cm^{3}} in dimension; only RGB and depth data covered by the cube is taken to generate a single patch.}{fig:patch_extraction}{0.45}

When all the synthetic images are generated and we have both real and synthetic data at hand, samples can be generated. For each of the images, we extract small patches covering and centered on the object. This is done by virtually setting a cube, of 40 \si{cm^{3}} in dimension, centered at the object's center of mass as shown in Fig.~\ref{fig:patch_extraction}. 
When all the patches are extracted, we normalize them. RGB channels are normalized to the zero mean and unit variance. The depth values within the defined bounding cube are normalized and mapped to the range $[0, 1]$ and the rest of the values are clipped. Finally, each patch $x$ is stored within a sample in addition to the object's identity $c$ and its pose $q$. The next step is to divide the samples between the sample sets $S_{train}$, $S_{db}$ and $S_{test}$, accordingly.



The template set $S_{db}$  contains only synthetic samples with the renderings coming from the coarse sampling (Fig.~\ref{fig:coarse}). It is used in both training (to form triplets) and test (as a database for the nearest neighbor search) phases. The samples of $S_{db}$ define a search database on which the nearest neighbor search is later performed. This is the main reason for coarse sampling: We want to minimize the size of the search database for faster retrieval. However, the sampling defined for the template set also directly limits the accuracy of the pose estimation.


\begin{figure}[t]
	\centering
	\begin{subfigure}[b]{0.51\linewidth}
		\centering
		\includegraphics[height=2.7cm]{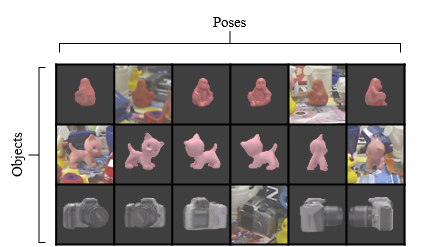}
		\caption{Training set $S_{train}$}
		\label{fig:traindb}
	\end{subfigure}\hfill
	\begin{subfigure}[b]{0.42\linewidth}
		\centering
		\includegraphics[height=2.7cm]{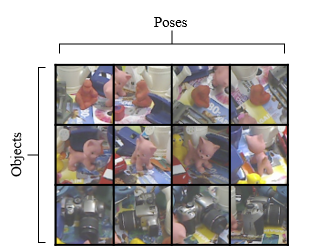}
		\caption{Test set $S_{test}$}
		\label{fig:testdb}
	\end{subfigure}
	\caption{Datasets: The training set $S_{train}$ consists of both real and synthetic (fine sampling); the test set $S_{test}$ consists of the real data not used for the training set $S_{train}$.}
	\label{fig:sets}
\end{figure}

The training set $S_{train}$ (Fig.~\ref{fig:traindb}) consists of a mix of synthetic and real data. The synthetic data represent samples coming from the renderings defined by the fine sampling (Fig.~\ref{fig:fine}). Approximately $50\%$ of the real data is added to the training set. This $50\%$ is selected by taking the real images that are close to the template samples in terms of the pose. The rest of the real samples are stored in the test set $S_{test}$ (Fig.~\ref{fig:testdb}), which is used to estimate the performance of the algorithm.

\subsection{Loss Function}
\begin{figure}[!htbp]
	\centering
	\begin{subfigure}[b]{.6\linewidth}
		\centering
		\includegraphics[height=2.7cm]{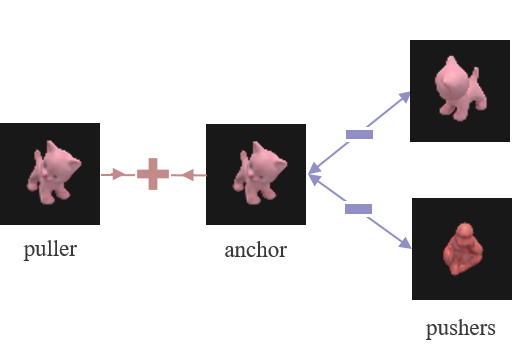}
		\caption{Triplet-wise term}
		\label{fig:triplet}
	\end{subfigure}\hfill
	\begin{subfigure}[b]{.4\linewidth}
		\centering
		\includegraphics[height=2.7cm]{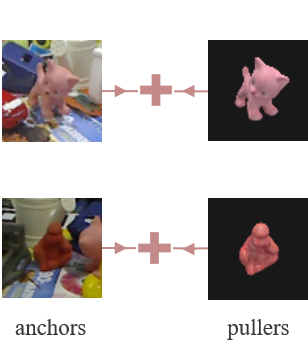}
		\caption{Pair-wise term}
		\label{fig:pair}
	\end{subfigure}
	\caption{CNN input format: triplets are used to learn a well-separated manifold, whereas pairs make the mapping invariant to various imaging conditions.}
	\label{fig:triplets_pairs}
\end{figure}

When the $S_{train}$ and $S_{db}$ sets are generated, we have all the data needed to start the training. The next step is to set the input format for the CNN, which is defined by its loss function. In our case, the loss function is defined as a sum of two separate loss terms $L_{triplets}$ and $L_{pairs}$:
\begin{eqnarray}
L = 
& L_{triplets} + L_{pairs}.
\end{eqnarray}
The first addend $L_{triplets}$ is a loss defined over a set $T$ of triplets, where a triplet is a group of samples $(s_i, s_j, s_k)$ selected such that $s_i$ and $s_j$ always come from the same object under a similar pose, and $s_k$ comes from either a different object or the same object under a less similar pose (Fig.~\ref{fig:triplet}). In other words, a single triplet consists of a pair of similar samples, $s_i$ and $s_j$, and a pair of dissimilar ones, $s_i$ and $s_k$. In our terminology, we call $s_i$ an anchor, $s_j$ a positive sample or a puller, and $s_k$ a negative sample or a pusher. The triplet loss component has the following form:
\begin{equation}
L_{triplets} = \sum_{(s_i,s_j,s_k) \in T} {max\left(0,1-\frac{||f(x_i)-f(x_k)||_2^2}{||f(x_i)-f(x_j)||_2^2+m}\right)},
\end{equation}
where $x$ is the input image of a certain sample, $f(x)$ is the output of the neural network given the input image, and $m$ is the margin, which introduces the margin for classification and sets the minimum ratio for the Euclidean distance of the similar and dissimilar pairs of samples. 

By minimizing $L_{triplets}$, one enforces two important properties that we are trying to achieve, namely: maximizing the Euclidean distance between descriptors from two different objects and setting the Euclidean distance between descriptors from the same object so that it is representative of the similarity between their poses.

The second addend $L_{pairs}$ is a pair-wise term. It is defined over a set $P$ of sample pairs $(s_i, s_j)$. Samples within a single pair come from the same object under either a very similar pose or the same pose but with different imaging conditions. Different imaging conditions may include illumination changes, different backgrounds, or clutter. It is also possible that one sample comes from the real data and the other from synthetic data. The goal of this term is to map two samples as close as possible to each other:

\begin{equation}
L_{pairs} = \sum_{(s_i,s_j) \in P} {||f(x_i)-f(x_j)||_2^2}.
\end{equation}

By minimizing the $L_{pairs}$, or the Euclidean distance between the descriptors, the network learns to treat the same object under different imaging conditions in the same way by mapping them onto the same point. Moreover, it ensures that samples with similar poses are set close together in the descriptor space, which is an important requirement for the triplet term.


\subsection{In-plane Rotations}
The initial method proposed in \cite{wohlhart2015learning} has a major limitation of not considering in-plane rotations or, in other words, omitting one additional degree of freedom. However, in real-world scenarios, it is hardly possible to omit in-plane rotations.

\insertfigure{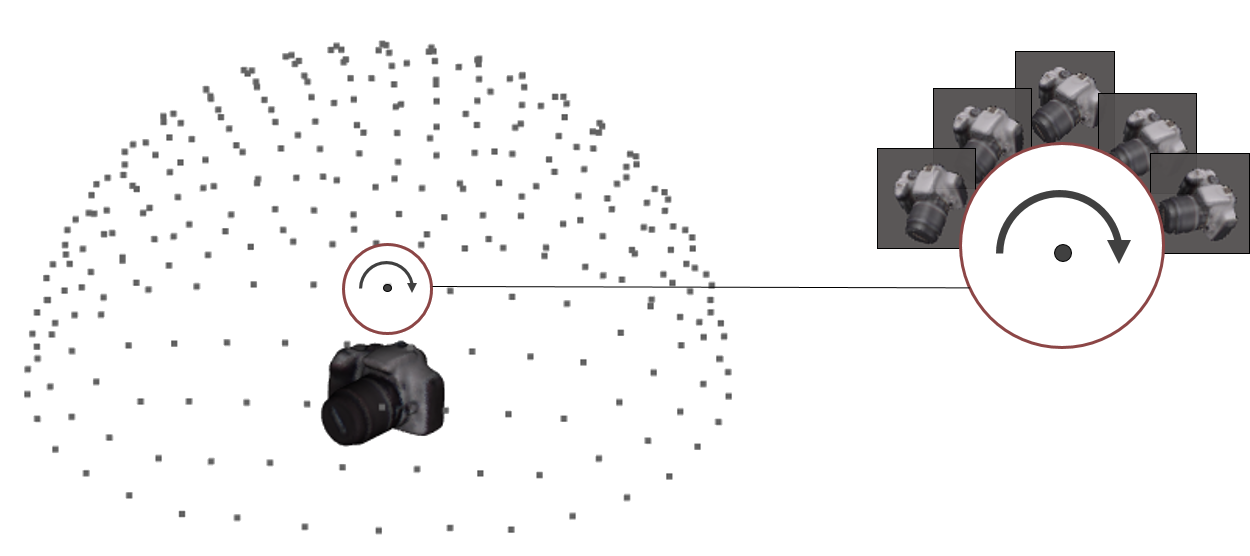}{In-plane rotations: At each vertex extra views are rendered by rotating the camera around the axis pointing at the object center.}{fig:inplane}{0.4}

In order to introduce in-plane rotations to the algorithm, one needs to generate additional samples with in-plane rotations and define a metric to compare the similarity between the samples in order to build triplets.

Generating synthetic in-plane rotated samples is relatively simple. What we need is to rotate the view camera at each sampling point (vertex) around its shooting axis and record a sample with a certain frequency as shown in Fig.~\ref{fig:inplane}. Currently, for the LineMOD dataset, we generate seven samples per vertex, going from -45 to 45 degrees with a stride of 15 degrees.

As for the similarity metric, we cannot use the dot product of the sampling point vectors anymore, as was proposed in the initial method, since we cannot incorporate an additional degree of freedom this way. Instead it was decided to use the quaternions $Q$ to represent rotations of the models and the angle between the samples' quaternions as a pose comparison metric $\theta(q_i, q_j) = 2 \arccos (|q_i \cdot q_j|)$.

\subsection{Triplet Loss with Dynamic Margin}
The triplet loss function, in the way it is used in \cite{wohlhart2015learning}, has one significant drawback. The margin term is a constant and is the same for all the different types of negative samples. This means that we are trying to push apart the objects of same and different classes with exactly the same margin term, whereas the desired goal is to map the objects of different classes farther away from each other. This slows down the training in terms of classification and results in a worse separation of the manifold. The logical solution to this is to set the margin term to be a variable and change it depending on the type of the negative sample.

\insertfigure{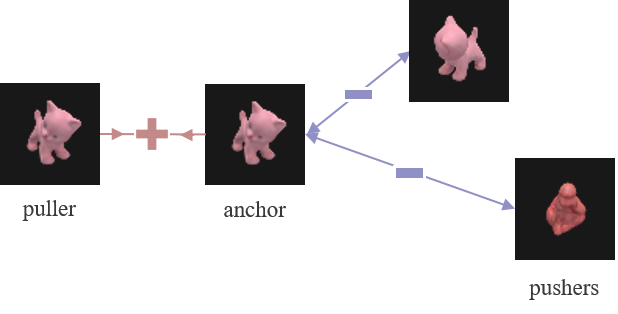}{Triplet loss with dynamic margin: better separation achieved by setting different inter- and intra-class margins.}{fig:tripletLossDM}{0.32}

We propose the following solution. If the negative sample belongs to the same class as the anchor, the margin term is set to be the angular distance between the samples. If, however, the negative sample belongs to a different class, the distance is set to a constant value that is larger than the maximal possible angular difference. The effect of the dynamic margin is illustrated in Fig~\ref{fig:tripletLossDM}. The updated loss function is defined as follows:
\begin{align}
&L_{triplets} = 
\sum_{(s_i,s_j,s_k) \in T} {max\left(0,1-\frac{||f(x_i)-f(x_k)||_2^2}{||f(x_i)-f(x_j)||_2^2+m}\right)}, \nonumber \\
&\text{where } m =
\begin{cases}
2 \arccos (|q_i \cdot q_j|) &\text{if } c_i = c_j,\\
n &\text{else, for } n > \pi.
\end{cases}
\end{align}

\subsection{Surface Normals}
Surface normals were considered an extra modality representing an object image, in addition to existing RGB and depth channels, to improve the algorithm accuracy. By definition, a surface normal defined at point $p$ is a 3D vector that is perpendicular to the tangent plane to the model surface at point $p$. Applied to many points on the model, surface normals result in a powerful modality describing its curvature.

In our pipeline, surface normals are calculated based on the depth map images (no additional sensor data required) using the method for the fast and robust estimation in dense range images proposed in work \cite{hinterstoisser2012gradient} and resulting in a \mbox{3-channel} modality. This approach allows smoothing of the surface noise and, therefore, allows for better surface normal estimates around depth discontinuities.

\comment{
This method considers the first order Taylor expansion of the depth function $D(x)$ around each pixel $x$ of the depth map:
\begin{equation}
D(x + dx) - D(x) = dx \nabla D + h.o.t.
\end{equation}
The neighboring pixels within a patch around $x$, defined by the offset term $dx$, are used to construct a system of equations that constrains the value of the depth gradient $\nabla D$. The optimal gradient $\hat{\nabla D}$, corresponding to a 3D plane, is then calculated by finding the least square solution of this system. Finally, the surface normal is estimated by computing the normalized cross product of the points defining the derived 3D plane.
}


\begin{table*}[h]
	\vspace{5pt}
	\centering
	\caption{Test setups: each underlined entry represents the tested parameter for a given test.}
	\label{tab:tests}
	\resizebox{0.98
		\textwidth}{!}{%
		\def\arraystretch{0.9}%
		\begin{tabular}{@{}ccccccccc@{}}
			\toprule
			& \textbf{Dataset} & \textbf{\begin{tabular}[c]{@{}c@{}}Training\\ data\end{tabular}} & \textbf{\begin{tabular}[c]{@{}c@{}}Testing\\ data\end{tabular}} & \textbf{\begin{tabular}[c]{@{}c@{}}In-plane\\ rotations\end{tabular}} & \textbf{\begin{tabular}[c]{@{}c@{}}Background\\ augmentation\end{tabular}} & \textbf{\begin{tabular}[c]{@{}c@{}}Data\\ channels\end{tabular}} & \textbf{\begin{tabular}[c]{@{}c@{}}Descriptor\\ dimension\end{tabular}} & \textbf{\begin{tabular}[c]{@{}c@{}}Triplet \\ margin type\end{tabular}} \\ \midrule
			\textbf{Test A} & \begin{tabular}[c]{@{}c@{}}LineMOD,\\ 15 objects\end{tabular} & synthetic + real & real & \begin{tabular}[c]{@{}c@{}}{}{\ul with and}\\ {\ul without}\end{tabular} & fractal noise & RGB-D & 32 & static \\
			\textbf{Test B} & \begin{tabular}[c]{@{}c@{}}LineMOD, \\ 6 objects\end{tabular} & synthetic + real & real & with & fractal noise & RGB-D & {\ul 3, 32} & \begin{tabular}[c]{@{}c@{}}{\ul static, dynamic}\end{tabular} \\
			\textbf{Test C} & \begin{tabular}[c]{@{}c@{}}LineMOD,\\ 15 objects\end{tabular} & {\ul synthetic} & real & with & \begin{tabular}[c]{@{}c@{}}{\ul white noise, fractal noise, }\\ {\ul random shapes, real backgrounds}\end{tabular} & RGB-D & 32 & dynamic \\
			\textbf{Test D} & \begin{tabular}[c]{@{}c@{}}LineMOD,\\ 15 objects\end{tabular} & synthetic + real & real & with & fractal noise & \begin{tabular}[c]{@{}c@{}}{\ul depth, normals,} \\ {\ul normals + depth}\end{tabular} & 32 & dynamic \\
			\textbf{Test E} & \begin{tabular}[c]{@{}c@{}}{ \ul BigBIRD,}\\ {\ul 50 objects}\end{tabular} & synthetic + real & real & without & fractal noise & RGB-D & 32 & {\ul static, dynamic} \\ \bottomrule
		\end{tabular}%
	}
	\vspace{-5pt}
\end{table*}

\subsection{Background Noise Generator}
\label{sec:background_noise}
One of the most difficult problems for computer vision methods is the treatment of clutter and different backgrounds in images. Since our samples do not have any background by default, it is difficult or sometimes impossible for the network to adapt to the real data full of noise and clutter in the background and foreground.

One of the easiest approaches to solving this problem is to use real images for training. Then, the network might adapt to the realistic data, but the major problem comes when no or very limited real data are available. In these cases, we have to teach the network to ignore the background or simulate the backgrounds as well as we can.

\begin{figure}[b] 
	\begin{subfigure}[b]{0.5\linewidth}
		\centering
		\includegraphics[width=0.75\linewidth]{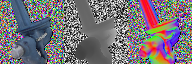}
		\caption{White noise} 
		\label{fig:noise_white} 
		\vspace{0.5ex}
	\end{subfigure}
	\begin{subfigure}[b]{0.5\linewidth}
		\centering
		\includegraphics[width=0.75\linewidth]{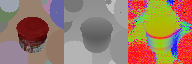} 
		\caption{Random shapes} 
		\label{fig:noise_shapes} 
		\vspace{0.5ex}
	\end{subfigure} 
	\begin{subfigure}[b]{0.5\linewidth}
		\centering
		\includegraphics[width=0.75\linewidth]{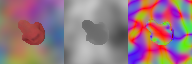} 
		\caption{Fractal noise} 
		\label{fig:noise_fractal} 
	\end{subfigure}
	\begin{subfigure}[b]{0.5\linewidth}
		\centering
		\includegraphics[width=0.75\linewidth]{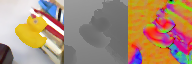} 
		\caption{Real backgrounds} 
		\label{fig:noise_real} 
	\end{subfigure} 
	\caption{Background noise types for synthetic data shown for different channels, i.e. RGB, depth, and normals.}
	\label{fig7} 
\end{figure}

In our implementation, we have a separate class generating different kinds of noise: white noise, random shapes, gradient noise, and real backgrounds.

The first and the simplest type of noise is white noise (Fig.~\ref{fig:noise_white}). To generate it, we simply sample a float value from 0 to 1 from a uniform distribution for each pixel. In the case of RGB, we do that three times for each pixel in order to fill all the channels.


The second type of noise is the random shape noise (Fig.~\ref{fig:noise_shapes}). The idea is to represent the background objects such that they have similar depth and color values. The color of the objects is again sampled from the uniform distribution, from 0 to 1, and the position is sampled from the uniform distribution, from 0 to the width of the sample image. This approach can also be used to represent foreground clutter by placing random shapes on top of the actual model.

The third type of noise we used is fractal noise (Fig.~\ref{fig:noise_fractal}), which is often used in computer graphics for texture or landscape generation and is the most advanced synthetic noise presented here. The fractal noise implementation we use is based on summing together multiple octaves of simplex noise first introduced by Ken Perlin in \cite{perlin2001noise}. It results in a smooth sequence of pseudo-random numbers avoiding rapid intensity changes, as in the case of white noise, which is much closer in spirit to the real-world scenarios.

The fourth and last type of noise is real backgrounds (Fig.~\ref{fig:noise_real}). Instead of generating the noise, we use \mbox{RGB-D} images of real backgrounds in a similar way to \cite{su2015render}. Given a real image, we randomly sample a patch of a needed size and use it as a background for a synthetically generated model. This noise modality is useful when we know beforehand in what kinds of environments the objects are going to be located.

One of the drawbacks of the baseline method is that the batches are generated and stored prior to execution. This means that at each epoch we use the same filled backgrounds over and over again, limiting the variability. To overcome this problem, in our implementation we generate batches online. At each iteration we fill the background of the chosen positive sample with one of the available noise modalities.

\section{Evaluation}

This section is devoted to the validation and evaluation of the implemented pipeline. After reproducing the results provided by the authors of the method \cite{wohlhart2015learning}, we performed a series of tests to evaluate the effect of the newly introduced modifications, e.g. in-plane rotations, surface normals, background noise types. Apart from that, we evaluated the performance of the algorithm on a larger dataset (BigBIRD). Note that all tests are preformed with the same network architecture as in \cite{wohlhart2015learning} for comparison reasons. For complete test setups refer to Table~\ref{tab:tests}.


\subsection{Tests on In-plane Rotations}
As we already know, the authors of the initially proposed method \cite{wohlhart2015learning} do not take in-plane rotations into account and do not include them in training, which is, however, needed for working in real-world scenarios. This test compares the performances of two networks: the one that is trained with in-plane rotations and the other that is trained without them. The goal is to see how avoiding in-plane rotations in training affects the performance on the test data with in-plane rotations and also to demonstrate the ability of the network to perform well with an additional degree of freedom introduced.

\begin{description}
	\item[\textit{Results:}]
	Given the setup, we compare the two above mentioned networks, labeled as baseline (without in-plane rotations) and baseline+ (with in-plane rotations), and obtain the results shown in Table~\ref{tab:inplane}.
	

	\begin{table}[h]
		\centering
		\caption{Comparison of the network trained without in-plane rotations (baseline) with the one trained using in-plane rotations (baseline+).}
		\label{tab:inplane}
		\resizebox{0.8\linewidth}{!}{%
			\def\arraystretch{1}%
			\begin{tabular}{@{}lcccc@{}}
				\toprule
				\multicolumn{1}{c}{\multirow{2}{*}{}} & \multicolumn{3}{c}{\textbf{Angular error}} & \multirow{2}{*}{\textbf{Classification}} \\ \cmidrule(lr){2-4}
				\multicolumn{1}{c}{} & 				\textbf{10$^{\circ}$} & \textbf{20$^{\circ}$} & \textbf{40$^{\circ}$} &   \\
				\midrule
				\textbf{Baseline} & 34.6\% & 63.8\% & 73.7\% & 81.9\% \\
				\textbf{Baseline+} & 60\% & 93.2\% & 97\% & 99.3\% \\ \bottomrule
			\end{tabular}
		}
	\end{table}
		
	The evaluation is performed only for a single nearest neighbor.
	As can be seen from Table~\ref{tab:inplane}, one gets a radical improvement over the results shown by the first modality, which is not trained to account for in-plane rotations. The results also demonstrate a successful adaptation to an additional degree of freedom.

\end{description}

\subsection{Tests on the Loss Function}
To evaluate the new loss function with dynamic margin, a set of tests comparing it with the old loss function was performed. Particularly, two tests were executed on six LineMOD objects (the lower amount is chosen for visualization purposes) using the best-performing training configurations for 3- and 32-dimensional output descriptors.

\begin{description}
	\item[Results:]

		\begin{figure}[b]
		\centering
		\begin{subfigure}[b]{.5\linewidth}
			\centering
			\includegraphics[height=2.63cm]{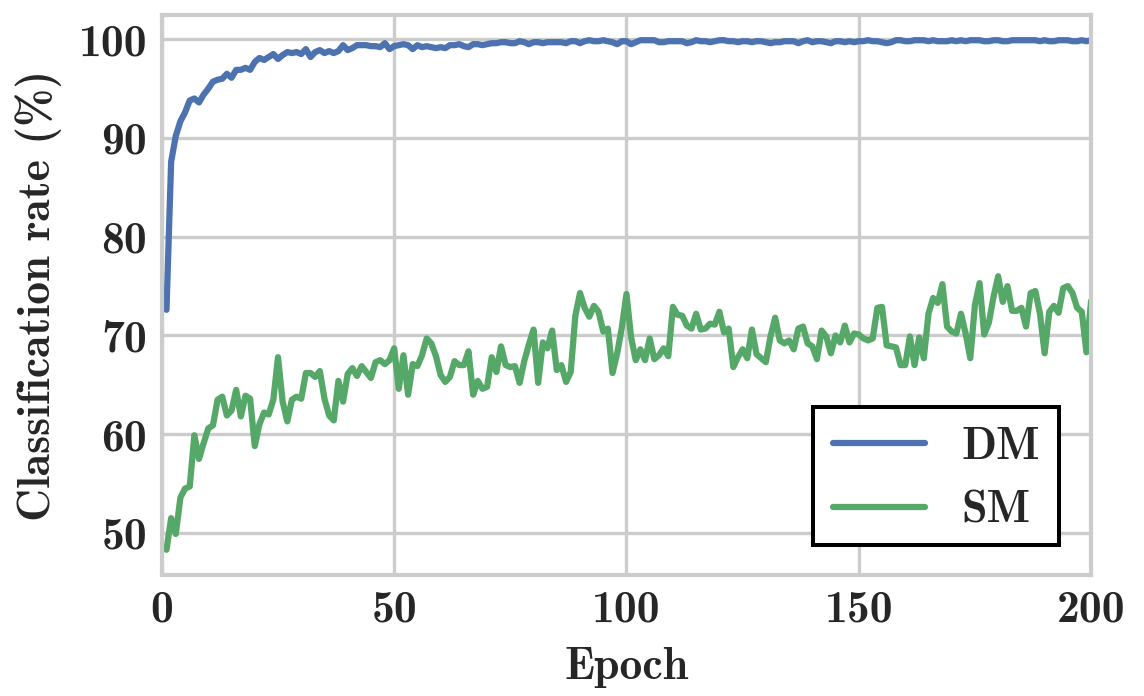}
			\caption{Classification rate}
			\label{fig:dm_3_class}
		\end{subfigure}%
		\begin{subfigure}[b]{.5\linewidth}
			\centering
			\includegraphics[height=2.63cm]{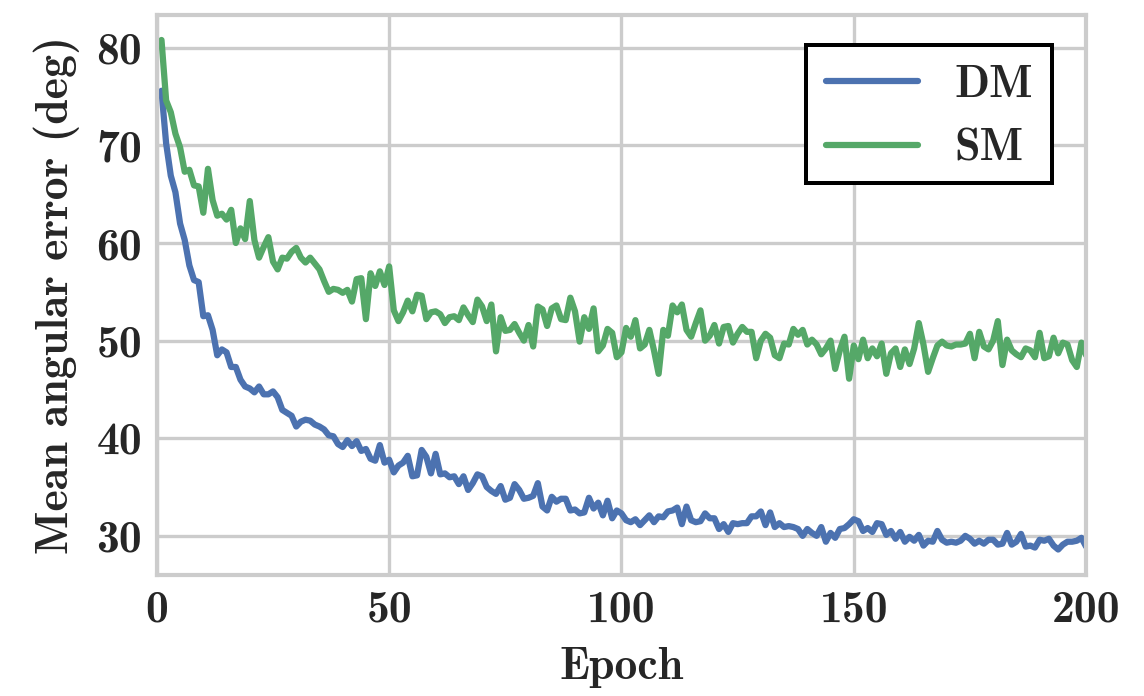}
			\caption{Mean angular error}
			\label{fig:dm_3_pose}
		\end{subfigure}
		\caption{Comparison of triplet loss with (DM) and without (SM) dynamic margin for the 3D output descriptor.}
		\label{fig:dm_3}
	\end{figure}

	Fig.~\ref{fig:dm_3} compares the classification rates (Fig.~\ref{fig:dm_3_class}) and mean angular errors for correctly classified samples (Fig.~\ref{fig:dm_3_pose}) over the set of training epochs (one run through the training set) for two modalities, i.e. the networks trained using the loss function with static and dynamic margins. It is clearly seen from the results that the new loss function makes a huge difference on the output result. It enables the network to learn a better classification much faster in comparison to the original. While the dynamic margin modality reaches 100\% classification accuracy very quickly, the older modality fluctuates around 70\%. Moreover, Fig.~\ref{fig:dm_3_pose} shows that we get a lower angular error for around 30\% more correctly classified samples.
	
		\begin{figure}[!h]
		\centering
		\begin{subfigure}[b]{.5\linewidth}
			\centering
			\includegraphics[width=0.8\linewidth]{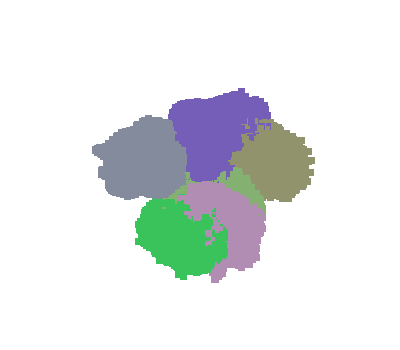}
			\caption{Static margin}
			\label{fig:sm}
		\end{subfigure}%
		\begin{subfigure}[b]{.5\linewidth}
			\centering
			\includegraphics[width=0.8\linewidth]{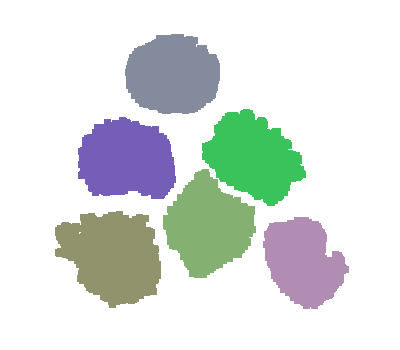}
			\caption{Dynamic margin}
			\label{fig:dm}
		\end{subfigure}
		\caption{Test set samples mapped to a 3D descriptor space: each color represents a separate object.}
		\label{fig:smdm}
	\end{figure}
	
	Fig. \ref{fig:smdm} shows the test samples mapped to the 3D descriptor space using the descriptor network trained with the old (Fig.~\ref{fig:sm}) and new (Fig.~\ref{fig:dm}) loss functions. The difference in the degree the objects are separated is explicit: in the right figure, the objects are well-separated preserving the minimal margin distance, resulting in a perfect classification score; the left figure still shows well-distinguishable object structures, but they are placed very close together and overlap, causing the classification confusion that is quantitatively estimated in Fig.~\ref{fig:dm_3_class}.
	
	In practice, however, we use dimensionally higher descriptor spaces, which improves both classification and angular accuracies. Fig.~\ref{fig:dm_32} shows the same charts as Fig.~\ref{fig:dm_3} but for a descriptor of a higher dimension, i.e. 32D. This results in a significant quality jump for both modalities, but the tendency stays the same: the new modality learns the classification much faster and provides a better angular accuracy for a larger number of correctly classified test samples.
	\begin{figure}[b]
		\centering
		\begin{subfigure}[b]{.5\linewidth}
			\centering
			\includegraphics[height=2.63cm]{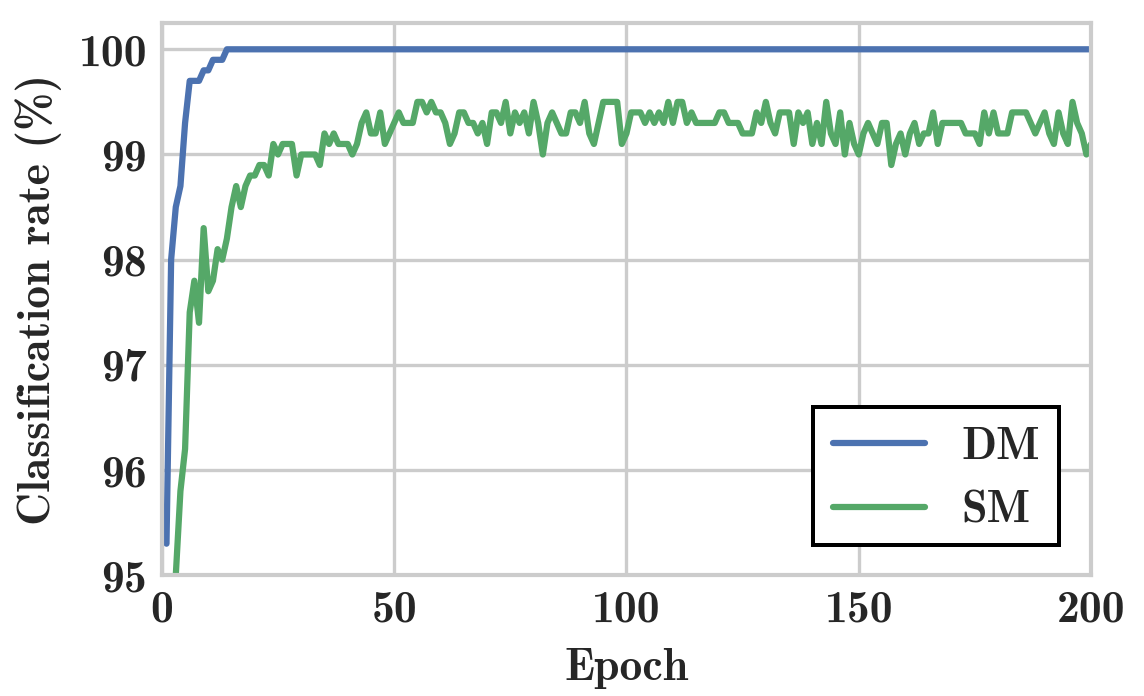}
			\caption{Classification rate}
			\label{fig:dm_32_class}
		\end{subfigure}%
		\begin{subfigure}[b]{.5\linewidth}
			\centering
			\includegraphics[height=2.63cm]{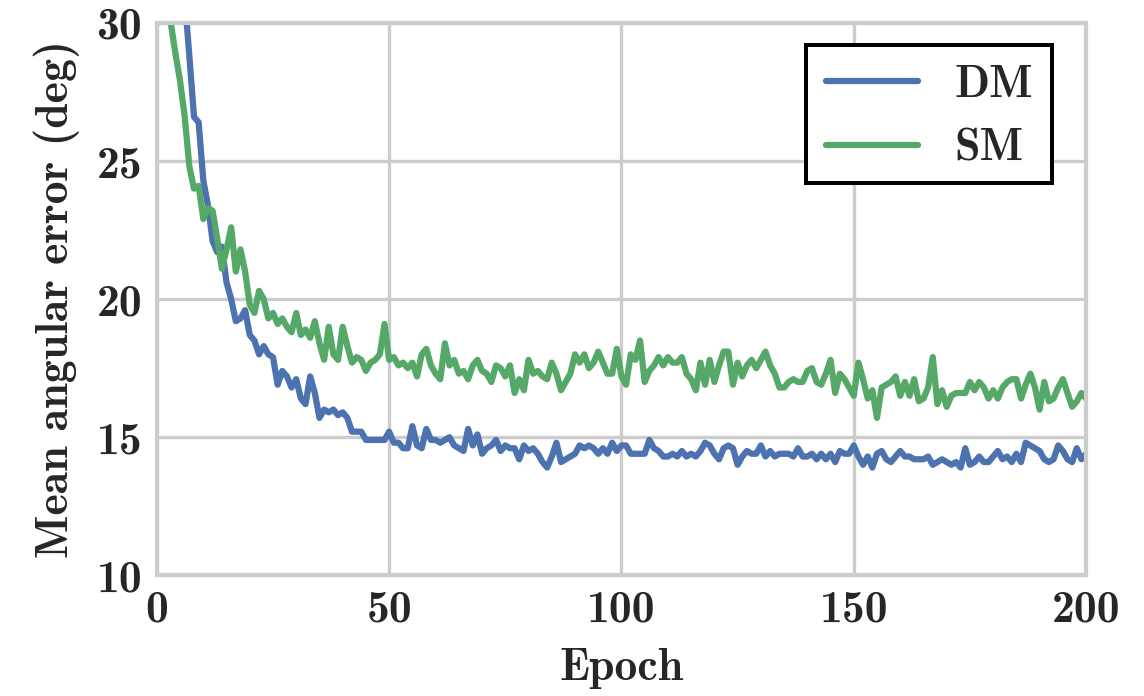}
			\caption{Mean angular error}
			\label{fig:dm_32_pose}
		\end{subfigure}
		\caption{Comparison of triplet loss with (DM) and without (SM) dynamic margin for 32D output descriptor.}
		\label{fig:dm_32}
	\end{figure}
	

\end{description}

\subsection{Tests on Background Noise Types}
Since we often do not have real \mbox{RGB-D} sequences on hand in real-world applications, but only 3D models provided, it would be beneficial to avoid using real data in training. The purpose of the following test is to show how well the network can adapt to the real data by only using the synthetic samples with artificially filled backgrounds in training. Specifically, we compare four different background noise modalities introduced in Section~\ref{sec:background_noise}: white noise, random shapes, fractal noise, and real backgrounds. 


\begin{description}
	\item[Results:]
	
	Fig.~\ref{fig:noise} shows the classification and pose accuracies for the four mentioned background noise modalities. The white noise modality shows the overall worst results, achieving around 21\% of classification accuracy (Fig.~\ref{fig:noise_class}), a marginal improvement over randomly sampling objects from a uniform distribution. 
	
		\begin{figure}[!h]
		\centering
		\begin{subfigure}[b]{.5\linewidth}
			\centering
			\includegraphics[height=2.63cm]{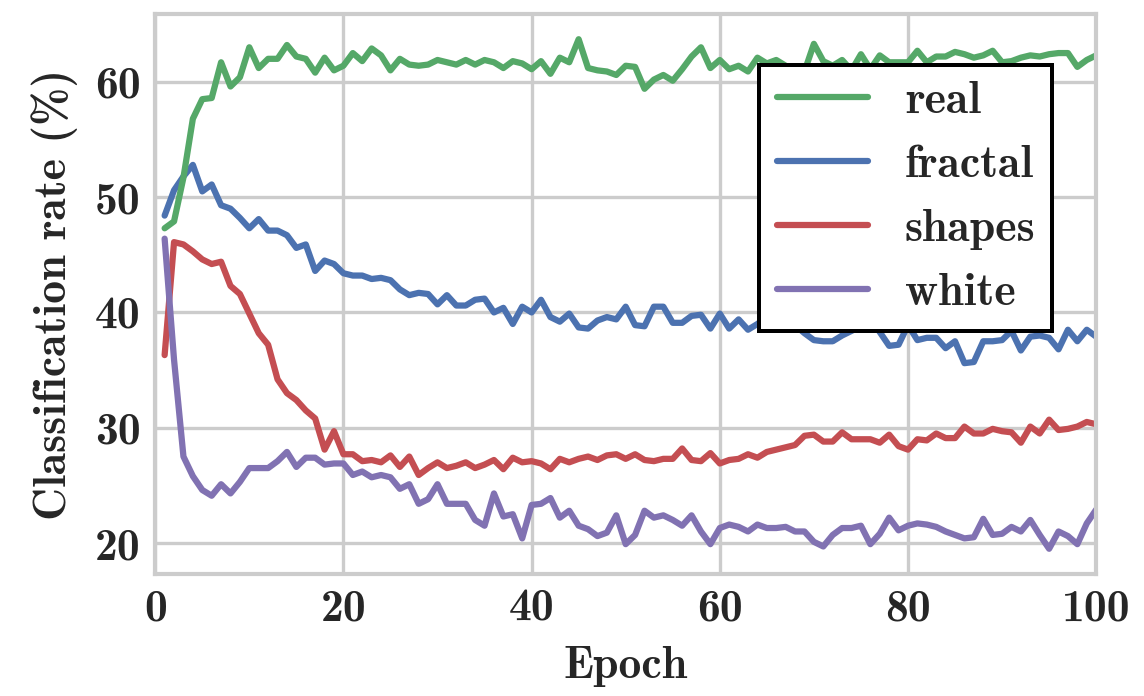}
			\caption{Classification rate}
			\label{fig:noise_class}
		\end{subfigure}%
		\begin{subfigure}[b]{.5\linewidth}
			\centering
			\includegraphics[height=2.63cm]{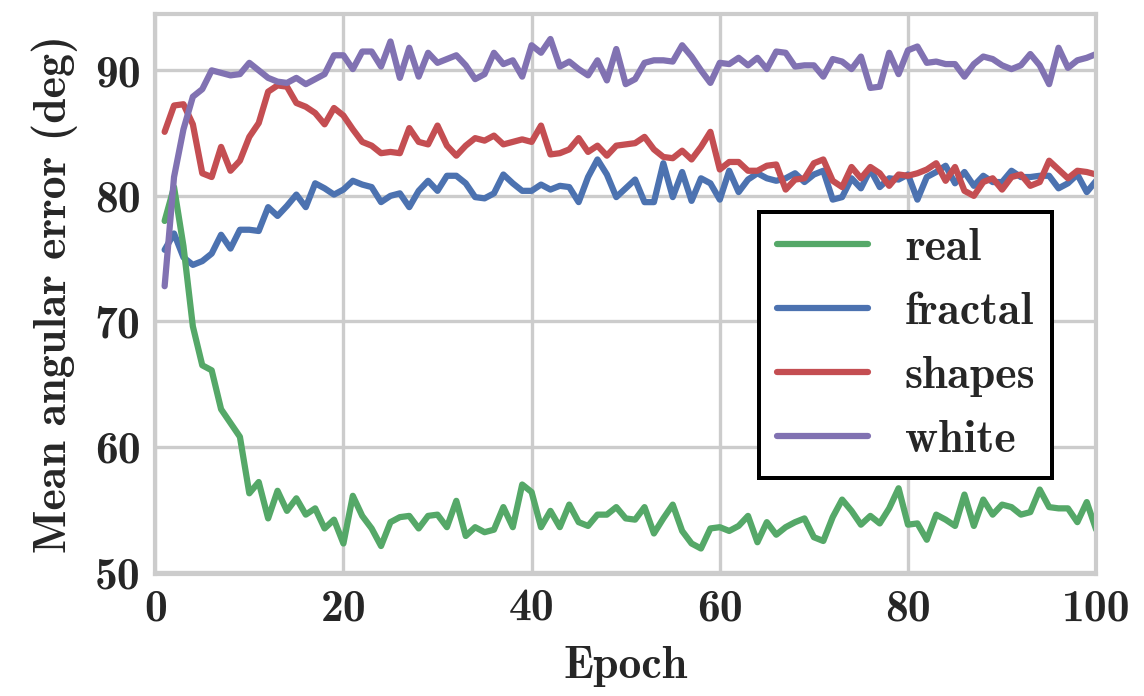}
			\caption{Mean angular error}
			\label{fig:noise_pose}
		\end{subfigure}
		\caption{Comparison of four different background noise modalities without any real data used for training.}
		\label{fig:noise}
	\end{figure}
	
	By switching to the random shapes modality, we get better results and fluctuate around 30\% of classification accuracy. The fractal noise modality shows the best results among the synthetic noise types and reaches up to 40\% of recognition rate. However, the real backgrounds modality outperforms fractal noise in classification terms and, moreover, shows better pose accuracy for a larger quantity of correctly classified samples (Fig.~\ref{fig:noise_pose}). As a result, if we can collect images from environments similar to the test set, the best option is to fill the backgrounds with the real images. If this is not the case or we do not have the environment specification beforehand, fractal noise is the preferred option.
	
\end{description}

\subsection{Tests on Input Image Channels}
In this test, we show the effect of the newly introduced normals channels. To do that, we demonstrate the influence of three input image channel modalities, i.e. depth, normals, and their combination on the output accuracy. More precisely, we use one of the most powerful pipeline configurations and train the network using the patches exclusively represented by the aforementioned channels.

\begin{description}
	\item[Results:]
	
	\begin{figure}[!b]
		\centering
		\begin{subfigure}[b]{.5\linewidth}
			\centering
			\includegraphics[height=2.63cm]{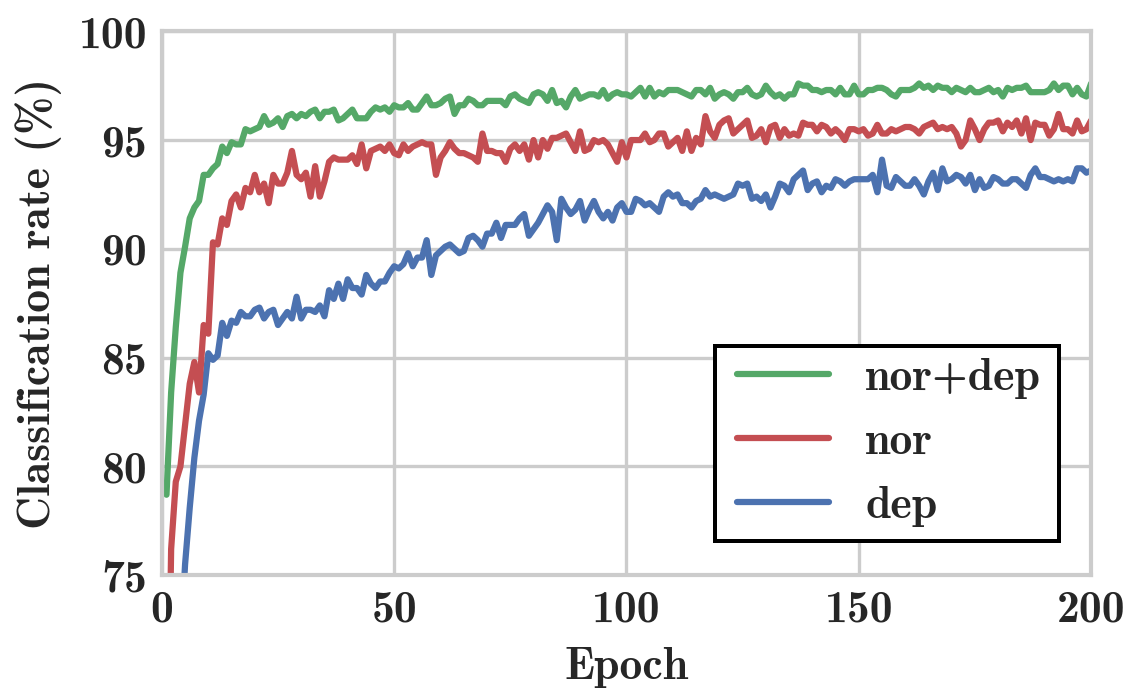}
			\caption{Classification rate}
			\label{fig:channels_class}
		\end{subfigure}%
		\begin{subfigure}[b]{.5\linewidth}
			\centering
			\includegraphics[height=2.63cm]{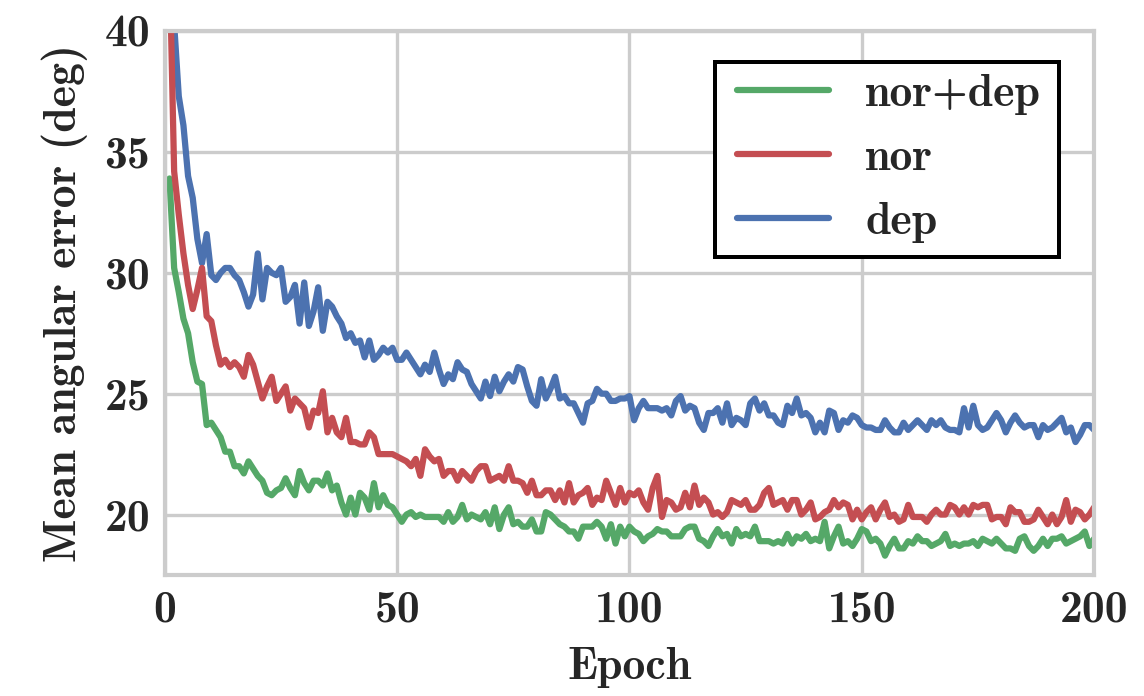}
			\caption{Mean angular error}
			\label{fig:channels_pose}
		\end{subfigure}
		\caption{Comparison of three modalities representing different input image channels used in training.}
		\label{fig:channels}
	\end{figure}

	Fig.~\ref{fig:channels} demonstrates the classification rate and pose error charts for three different networks trained on three different combinations of input patch channels: depth, normals, and normals+depth. It can be observed that the network trained on the newly introduced surface normals modality performs better than the one trained on the depth maps only. This is beneficial since surface normals are generated entirely based on the depth maps and no additional sensor data is needed. Additionally, by combining the surface normals and depth channels into a single modality, we get even better results compared to using them separately. More importantly, the same effect holds true when RGB channels are added to the presented modalities.
	
\end{description}

\subsection{Tests on Larger Datasets}
The goal of this experiment is to see how well the algorithm generalizes to a larger number of models. In particular, we want to evaluate how the increased amount of models in training affects the overall performance. Since the LineMOD dataset has only 15 models available, the adapted BigBIRD dataset, which offers many more models, is used for this test. 
\begin{description}
	
	\item[Results:]
	Given one of the most powerful pipeline configurations, we have trained the network on 50 models of the BigBIRD dataset. After finishing the training, we achieved the results shown in Table~\ref{tab:bigbird}.
	


\begin{table}[h]
	\caption{Angular error histogram computed using the samples of the test set for a single nearest neighbor.}
	\label{tab:bigbird}
	\centering
	\resizebox{0.8\linewidth}{!}{%
		\def\arraystretch{1}%
		\begin{tabular}{@{}lcccc@{}}
			\toprule
			\multicolumn{1}{c}{\multirow{2}{*}{}} & \multicolumn{3}{c}{\textbf{Angular error}} & \multirow{2}{*}{\textbf{Classification}} \\ \cmidrule(lr){2-4}
			\multicolumn{1}{c}{} & 				\textbf{10$^{\circ}$} & \textbf{20$^{\circ}$} & \textbf{40$^{\circ}$} &   \\
			\midrule
			\textbf{SM} & 67.4\% & 79.6\% & 83.5\% & 85.4\% \\
			\textbf{DM} & 67.7\% & 91.2\% & 95.6\% & 98.7\% \\
			\bottomrule
		\end{tabular}
	}
\end{table}
	
	Table~\ref{tab:bigbird} shows the histogram of classified test samples for several tolerated angle errors. The results are encouraging: for 50 models each represented by approximately 300 test samples, we get a classification of 98.7\% and a very good angular accuracy, the significant improvement over the old loss function. As a result, this approach proves to scale well with respect to the number of models, making it suitable for industrial applications.

	

	
\end{description}

\section{CONCLUSIONS}


In this work, the method first introduced in \cite{wohlhart2015learning} was improved in terms of its learning speed, robustness to clutter, and usability in real-world scenarios. We have implemented a new loss function with dynamic margin that allows for a faster training and better accuracy. Furthermore, we introduced in-plane rotations (present in real-world scenarios) and new background noise types (to better mimic the real environments), added surface normals as an additional powerful image modality, and created an efficient method to generate batches allowing for higher variability during training.


\addtolength{\textheight}{-12cm}   



%
%
%


\bibliography{literature}

\end{document}